%% file: root.tex
\documentclass[letterpaper, 10 pt, conference]{ieeeconf}  

\IEEEoverridecommandlockouts                              

\usepackage{graphics} 
\usepackage{epsfig} 

\usepackage{amsmath} 
\usepackage{amssymb}  
\usepackage{multirow}
\usepackage{booktabs}
\usepackage{graphicx}
\usepackage{subcaption}
\usepackage{tikz}

\usepackage[dvipsnames]{xcolor}
\usepackage[table]{xcolor}
\usepackage{algpseudocode} 
\usepackage[ruled,vlined]{algorithm2e} 

\definecolor{citypink}{RGB}{227, 108, 194}
\definecolor{cityblue}{RGB}{128, 159, 225}
\definecolor{SunsetOrange}{RGB}{217, 123, 41}
\definecolor{AmberGold}{RGB}{255, 191, 0}
\definecolor{NeonOrange}{RGB}{255, 95, 31}
\definecolor{FluoroYellow}{RGB}{204, 255, 0}

\usepackage{indentfirst}
\usepackage{titletoc}
\usepackage{cuted}
\usepackage{capt-of}
\usepackage{gensymb}
\usepackage{courier}
\usepackage{cleveref}
\usepackage{multirow}

\usepackage{url}
\usepackage{balance}

\definecolor{colorbest}{RGB}{255,179,179}
\definecolor{colorsecond}{RGB}{255,217,179}
\definecolor{colorthird}{RGB}{255,255,179}
\newcommand{\best}[0]{\cellcolor{colorbest} }
\newcommand{\second}[0]{\cellcolor{colorsecond}}

\usepackage{cite}

\def\paper{OPFA}

\title{\vspace{-2mm}
\LARGE \bf
\raisebox{-0.01\linewidth}{\includegraphics[width=0.8cm]{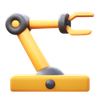}} 
    \textsf{\textcolor{AmberGold}{One-Policy-Fits-All}}: 
    Geometry-Aware Action Latents for \\ Cross-Embodiment Manipulation
}

\author{
\textbf{Juncheng Mu}$^{1,2*}$ \quad 
\textbf{Sizhe Yang}$^{1,3*}$ \quad 
\textbf{Hojin Bae}$^{2*}$ \quad 
\textbf{Feiyu Jia}$^{1,4}$ \\
\textbf{Qingwei Ben}$^{1,3}$ \quad 
\textbf{Boyi Li}$^{5\dagger}$ \quad 
\textbf{Huazhe Xu}$^{2\dagger}$ \quad 
\textbf{Jiangmiao Pang}$^{1\dagger}$ \\
$^{1}$Shanghai AI Laboratory \quad
$^{2}$Tsinghua University \quad
$^{3}$The Chinese University of Hong Kong \\
$^{4}$University of Science and Technology of China \quad
$^{5}$NVIDIA \\
Project page: \textcolor[HTML]{D81B60}{\url{https://mujc2021.github.io/opfa/}}
}

\begin{document}

\maketitle
\thispagestyle{empty}
\pagestyle{empty}

\input{sec/abstract}
\input{sec/intro}

\input{figures/pipeline}

\input{sec/related_works}

\input{sec/method}

\input{sec/experiments}
\input{sec/conclusion}






\vspace{1mm}
{
    \bibliographystyle{IEEEtran}
    \bibliography{IEEEabrv}
}

\end{document}

%% file: sec/abstract.tex
\input{figures/teaser}
\begin{abstract}

Cross-embodiment manipulation is crucial for enhancing the scalability of robot manipulation and reducing the high cost of data collection.
However, the significant differences between embodiments, such as variations in action spaces and structural disparities, pose challenges for joint training across multiple sources of data. To address this, we propose \textit{One-Policy-Fits-All} (\paper), 
a framework that enables learning a single, versatile policy across multiple embodiments.
We first learn a \textit{Geometry-Aware Latent Representation} (GaLR), which leverages 3D convolution networks and transformers to build a shared latent action space across different embodiments.
Then we design a unified latent retargeting decoder that extracts embodiment-specific actions from the latent representations, without any embodiment-specific decoder tuning.
\paper\ enables end-to-end co-training of data from diverse embodiments, including various grippers and dexterous hands with arbitrary degrees of freedom, significantly improving data efficiency and reducing the cost of skill transfer. We conduct extensive experiments across 11 different end-effectors. 
The results demonstrate that \paper\ significantly improves policy performance in diverse settings by leveraging heterogeneous embodiment data.
For instance, cross-embodiment co-training can improve success rates by more than 50\% compared to single-source training.
Moreover, by adding only a few demonstrations from a new embodiment (e.g., eight), \paper\ can achieve performance comparable to that of a well-trained model with 72 demonstrations.

\end{abstract}

%% file: figures/teaser.tex
\begin{strip}
\centering
\vspace{-15mm}
\includegraphics[width=\textwidth]{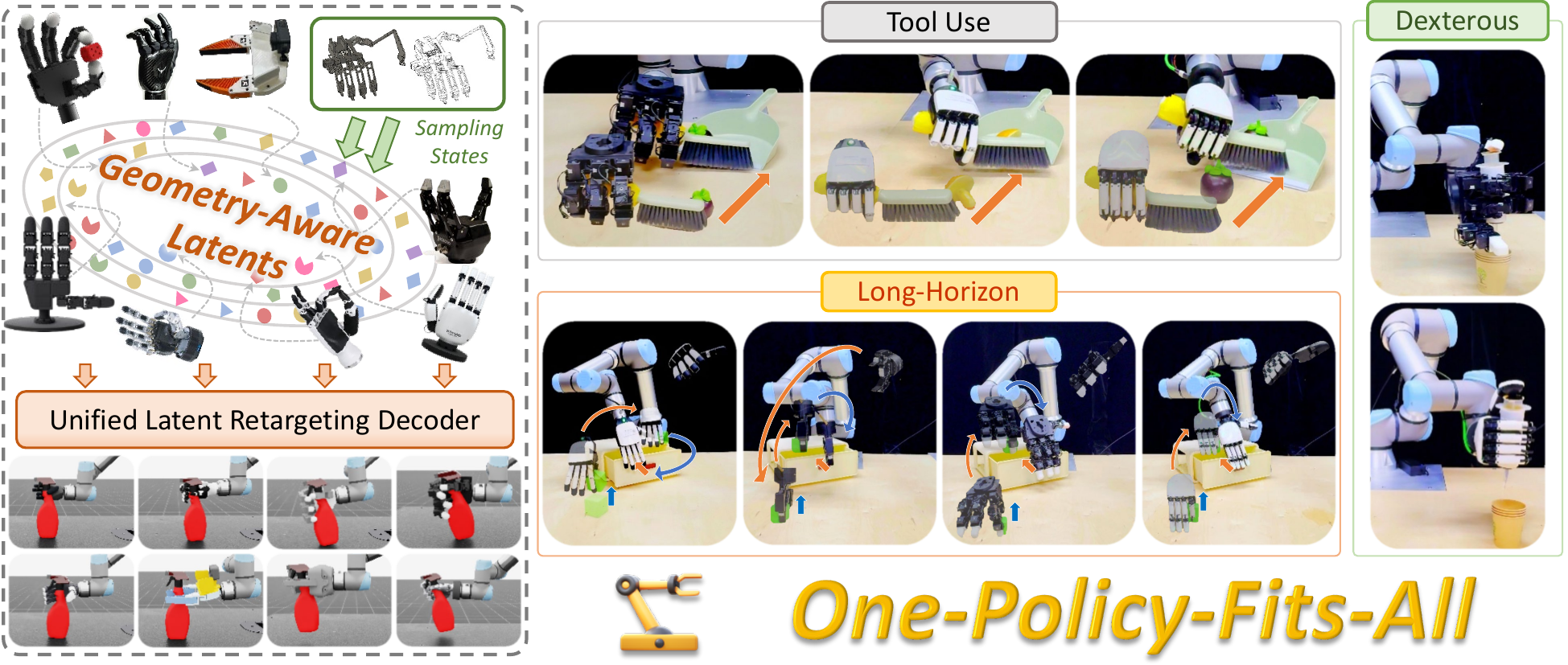}
\vspace{-6mm}
\captionof{figure}{We introduce \textbf{One-Policy-Fits-All} (OPFA), a general framework for cross-embodiment manipulation. OPFA leverages the geometric structures of diverse end-effectors to construct a unified latent action representation, and employs a unified latent retargeting decoder to recover embodiment-specific actions. This design enables seamless skill transfer across grippers and dexterous hands, offering a scalable solution to data scarcity and enabling rapid adaptation to new embodiments.}
\vspace{-4mm}
\label{fig:teaser}
\end{strip}

%% file: sec/intro.tex
\section{INTRODUCTION}
\vspace{-1mm}

Imitation learning \cite{behavior_clone, deepmimic, GAIL} has emerged as a pivotal paradigm in robotics, offering learning-based solutions for complex real-world manipulation tasks. However, unlike learning in text and vision domains \cite{llama, llava, qwen} where data typically comes from a single modality, robotic manipulation is inherently coupled with the physical characteristics of the end-effector, posing two key challenges for policy generalization:
(i) end-effectors vary drastically in morphology and degrees of freedom—from simple grippers to highly articulated anthropomorphic hands, resulting in fundamentally different action spaces that hinder the joint exploitation of cross-embodiment data;
(ii) introducing a new end-effector typically requires collecting a large amount of embodiment-specific data, which is both costly and limits scalability.

Existing methods struggle to fully exploit cross-embodiment data due to two main challenges: (i) substantial differences in action dimensions and spaces across embodiments, and (ii) large structural discrepancies that create a pronounced embodiment gap.
If a unified policy could co-train cross-embodiment data without conflicts, the problem of data scarcity would be effectively alleviated.
Furthermore, decoupling policy learning from a specific end-effector would enable more versatile and generalizable policies, as well as more data-efficient learning, thereby greatly facilitating practical real-world deployment.

Recently, several approaches \cite{rdt1b, hrdt, pi0, pi05, octo, dexvla} have explored cross-embodiment co-training to expand data scale and improve generalization across embodiments. Most of these methods target settings where grippers act as the end-effector, achieving cross-embodiment capability at the arm–gripper level either by unifying the action space \cite{rdt1b} or by employing embodiment-specific decoders \cite{octo}. However, extending such strategies to dexterous hands—with their higher degrees of freedom and more complex physical structures—remains substantially more challenging.
Naively assigning a separate action head or decoder to each embodiment \cite{dexvla, latent_action_diffusion} neglects the geometric structure of the hand, while restricting each decoder to single-source data, leading to low data efficiency and limited generalization.

To overcome the limitations of prior methods and enable cross-embodiment co-training across a wide range of end-effectors, particularly dexterous hands, we propose One-Policy-Fits-All (OPFA), a general framework for cross-embodiment manipulation. Specifically, OPFA first learns a geometry-aware action latent representation (GaLR), which captures the spatial structure of an end-effector’s reachable states from point clouds (derived directly from joint angles) using 3D convolutional networks \cite{kpconv} and transformers \cite{geotransformer}.
The GaLR produced by the encoder unifies action space dimensions across multiple embodiments while consistently encoding geometric information. This transforms action prediction into a latent-space prediction of spatial structures across different embodiments.
During training, reachable-state point clouds from multiple embodiments are sampled, and the decoder recovers embodiment-specific joint angles from the encoded GaLR. This enables the encoder and decoder to be trained jointly in an end-to-end manner without any additional manual annotation.
It is noteworthy that, unlike prior approaches \cite{dexvla, latent_action_diffusion, octo} that train separate decoders for each embodiment, we design a unified latent retargeting decoder capable of handling diverse embodiments.
Finally, the trained spatial encoder, decoder, and the constructed GaLR can be integrated into diverse action prediction methods (e.g., DP \cite{dp}, DP3 \cite{dp3}), effectively transferring action prediction to the latent space. In this way, data from multiple embodiments are represented in a unified dimensionality while sharing geometric information, substantially enhancing generalization across diverse end-effectors.

Extensive experiments in both simulation and the real world validate the strong performance of OPFA. We evaluate it on 11 diverse embodiments—ranging from two-finger grippers (UMI \cite{umi}, Robotiq-2F), to three-finger (Robotiq-3F), four-finger (Leap \cite{leaphand}, Allegro), and five-finger hands (Inspire Hand \cite{inspirehand}, XHand \cite{xhand}, etc.)—across 14 challenging manipulation tasks such as pouring and sweeping. Comprehensive cross-embodiment evaluations show that OPFA consistently outperforms both single-embodiment training and naive cross-embodiment co-training with embodiment-specific decoders. Notably, OPFA can achieve success rates on a new end-effector comparable to those of a well-trained model (72 demonstrations), while requiring as few as eight demonstrations.
In summary, our contributions are three-fold:
\begin{itemize}
    \item We construct a Geometry-Aware Latent Representation (GaLR) that aligns the action space dimensions across different end-effectors and learns geometric information without any manual annotation cost.
    \item We propose an end-to-end cross-embodiment co-training method that does not require any embodiment-specific decoder tuning.
    \item We conduct extensive experiments in both simulation and the real world, covering 11 different end-effectors, and demonstrate that OPFA significantly outperforms both self-training and naive co-training methods in various cross-embodiment settings.
\end{itemize}

%% file: figures/pipeline.tex
\begin{figure*}
\centering
\vspace{-1mm}
\includegraphics[width=\textwidth]{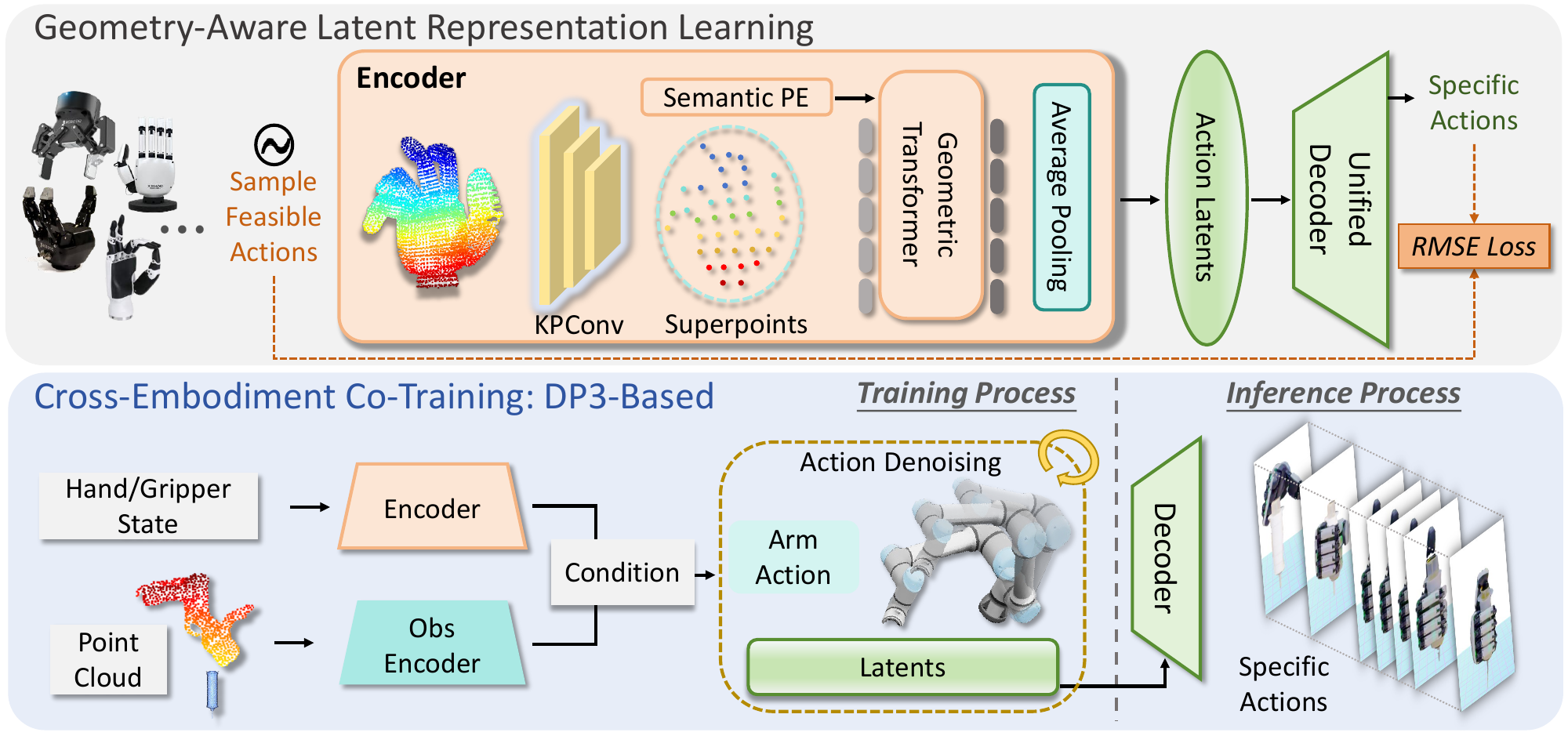}
\caption{The training pipeline of OPFA follows a two-stage paradigm.
(1) We first construct a \textit{Geometry-Aware Latent Representation} (GaLR) by encoding sampled reachable-state point clouds with 3D convolutions and geometric transformers for local/global feature extraction. A unified latent retargeting decoder then disentangles embodiment-specific actions from the latent space, enabling end-to-end training without manual annotations.
(2) The pretrained encoder–decoder pair is integrated into any downstream policy (e.g., DP3), allowing cross-embodiment data to be jointly trained in a unified latent action space.}
\label{fig:pipeline}
\vspace{-6mm}
\end{figure*}

%% file: sec/related_works.tex
\section{RELATED WORKS}





\subsection{Cross-Embodiment Learning for Robot Arms}

Recent efforts toward a universal policy have focused on cross-embodiment learning, with significant advancements for robot arms equipped with parallel-jaw grippers. These policies are often trained on large-scale datasets \cite{Jacquard,bridgeData,ACRONYM,RH20T,QT-Opt,Hand-Eye, RobotsHome,BridgeDataV2,rt-x}. To handle the data heterogeneity across different robots, a common strategy is to unify the action space around the end-effector. 
Models like RT-1 \cite{rt1} and RT-2 \cite{rt2} predict end-effector motions and gripper states from visual and language inputs.
Octo \cite{octo} utilizes a standardized end-effector space for pre-training and is uniquely designed to be adapted to new action spaces via modular adapters during fine-tuning.
A more comprehensive approach is taken by RDT-1B \cite{rdt1b}, which standardizes all data into the \textit{Physically Interpretable Unified Action Space}.
Another effective paradigm involves learning a shared latent space to bridge morphological gaps between different arms\cite{wang2024crossembodimentrobotmanipulationskill}. These methods have proven effective for robot arms, which typically use simple parallel-jaw grippers. However, they do not directly address the distinct challenges posed by dexterous hands, which feature complex geometries and high degrees of freedom.


\subsection{Cross-Embodiment Learning for Dexterous Hands}
Cross-embodiment learning for multifingered robotic hands \cite{adagrasp,unigrasp} is more challenging due to their high dimensionality and significant structural diversity. A common strategy involves training embodiment-specific decoders \cite{dexvla, latent_action_diffusion}. However, these approaches still restrict each decoder to training on data from a single embodiment.
Other works leverage human motion as a prior via retargeting.
Bauer et al. \cite{latent_action_diffusion} trains a diffusion policy in a shared latent action space, with MANO-based \cite{MANO} latent feature aligning.
VideoDex \cite{videodex} extracts actions directly from human videos. CrossDex \cite{crossdex} learns a universal policy in an abstract action space based on MANO model \cite{MANO}, which is then retargeted to robot-specific commands.
However, due to the substantial structural discrepancies across different end-effectors, naively aligning them to a human-hand model may induce action conflicts.
Instead of relying on pre-defined human hand priors, OPFA directly learns a Geometry-Aware Latent Representation based on the sampled reachable-state point cloud of all the end-effectors to avoid action conflicts, and utilizes a unified decoder based on latent retargeting without any embodiment-specific decoder tuning.

%% file: sec/method.tex
\section{METHOD}

We propose \textit{One-Policy-Fits-All} (OPFA), a general framework for cross-embodiment manipulation. OPFA adopts a two-stage paradigm: (i) a learning-based but annotation-free stage that learns a \textit{Geometry-Aware Latent Representation} (GaLR), unifying the action spaces of diverse end-effectors; and (ii) a policy integration stage, where GaLR is embedded into the state condition and action prediction head of the policy (e.g., DP3 \cite{dp3}), enabling end-to-end cross-embodiment co-training. The overall pipeline is shown in \Cref{fig:pipeline}.

\subsection{Overview}

We denote the set of embodiments as $\mathcal{M} = \{1,2,\dots,M\}$,
where each embodiment \(m \in \mathcal{M}\) is associated with a dataset $\mathcal{D}_m = \{\tau^m_i\}_{i=1}^{N_m}$, consisting of collected trajectories. Each trajectory is defined as
$\tau^m_i = \big\{ \big(\mathbf{o}^m_t, \mathbf{a}^m_t \big) \big\}_{t=1}^{T^m_i}$,
where \(\mathbf{o}^m_t\) denotes the observation at timestep \(t\),
\(\mathbf{a}^m_t \in \mathbb{R}^{d_m}\) is the corresponding action,
and \(T^m_i\) is the length of the trajectory.
To fully leverage data from multiple embodiments, we aim to co-train $\{\mathcal{D}_m\}_{m=1}^M$. 
However, since $d_m$ differs for each embodiment, unifying the action space presents a major challenge. 
Prior works \cite{latent_action_diffusion, crossformer} often address this issue by applying embodiment-specific transformer heads or action decoders. 
However, for embodiment $m$, the embodiment-specific decoder can only be trained with data from $\mathcal{D}_m$.
This limitation becomes particularly severe in few-shot learning scenarios, where data scarcity causes decoders to overfit and hinders effective skill transfer (see \Cref{sec:few-shot learning}).

\subsection{Learning Geometry-Aware Latent Representation}

The construction of GaLR adopts an end-to-end encoder–decoder training scheme, with all training data generated automatically, entirely \textit{without manual annotation}.
For each embodiment $m \in \mathcal{M}$, we first sample a set of reachable states 
$J^m$, and apply the forward kinematics function $f_{\text{FK}}^m$ and point
sampling to obtain a set of training data $\{ (\mathbf{a^m}, \mathcal{P}) \}^m$, where $\mathbf{a^m} \in J^m$ is the joint angles used for supervision,
and $\mathcal{P} \in \mathbb{R}^{|\mathcal{P}| \times 3}$ is the sampled point cloud. 
We apply a vision encoder $f_\theta$ to extract shared geometric features across 
multiple embodiments and states, mapping each $\mathcal{P}$ into the latent space: $\mathbf{z} = f_\theta(\mathcal{P}) \in \mathbb{R}^{d_{latent}}$.
A unified decoder $g_\psi$ then predicts the embodiment-specific joint angles 
$\mathbf{\hat a^m} \in \mathbb{R}^{d_m}$ from the latent vector: $\mathbf{\hat a^m} = g_\psi(\mathbf{z})$.

Because the number of dense points is relatively large, directly extracting features at this level results in redundancy and low computational efficiency. To address this, we perform three stages of downsampling to obtain multi-scale spatial features. Specifically, we denote the original dense point cloud as $\mathcal{P} \in \mathbb{R}^{|\mathcal{P}| \times 3}$, the first-level downsampled points as $\Tilde{\mathcal{P}} \in \mathbb{R}^{|\Tilde{\mathcal{P}}| \times 3}$, and the final downsampled points (\textit{superpoints}) as $\Hat{\mathcal{P}} \in \mathbb{R}^{|\Hat{\mathcal{P}}| \times 3}$, where $|\Hat{\mathcal{P}}| < |\Tilde{\mathcal{P}}| < |\mathcal{P}|$. We then apply a geometric transformer \cite{geotransformer} at the superpoint level $\Hat{\mathcal{P}}$ to capture global gesture-aware representations.

\noindent
\textbf{Multi-scale local structure encoding.}
We perform multi-scale feature extraction on the downsampled point clouds with the 3D convolution \cite{kpconv}.
Let $\mathcal{B}^3_r=\{\mathbf{y}\in \mathbb{R}^3\ |\ \Vert \mathbf{y} \Vert \le r \}$ be a ball of radius $r$ 
centered at a query point $\mathbf{x}$. For a neighboring point $\mathbf{x_i}$, let $\mathbf{y_i} = \mathbf{x_i} - \mathbf{x} \in \mathcal{B}^3_r$ denote the relative coordinate of $\mathbf{x_i}$. The kernel points are defined as 
$\{\mathbf{\Tilde{x}_k} \mid k < K\} \subset \mathcal{B}^3_r$, 
and each kernel point $\mathbf{\Tilde{x}_k}$ is associated with a weight matrix 
$\mathbf{W}_k \in \mathbb{R}^{D_{\mathrm{in}} \times D_{\mathrm{out}}}$, 
where $K$ is the number of kernel points, and $D_{\mathrm{in}},D_{\mathrm{out}}$ 
are the input and output feature dimensions. 

 For a query point $\mathbf{x}$ with neighborhood $\{\mathbf{x_i}\}$ and input features $f_{\mathbf{x_i}} \in \mathbb{R}^{D_{\mathrm{in}}}$, 
the convolution output is given by
\begin{equation}
    g(\mathbf{x}) = \sum_{i} \sum_{k<K} h(\mathbf{y_i},\mathbf{\Tilde{x}_k})\,\mathbf{W}_k f_{\mathbf{x_i}},
\end{equation}
where $h$ measures the proximity between $\mathbf{y_i}$ and $\mathbf{\Tilde{x}_k}$. $h$ is implemented as a linear decay function truncated at $\sigma$: 
\begin{equation}
    h(\mathbf{y_i},\mathbf{\Tilde{x}_k}) = \max\!\left(0, 1 - \frac{\Vert \mathbf{y_i} - \mathbf{\Tilde{x}_k} \Vert}{\sigma}\right).
\end{equation}
Here, $\sigma$ is a hyperparameter that controls the influence radius of each kernel point.

\noindent
\textbf{Global gesture perception.}
The 3D convolution stage effectively extracts multi-scale local geometric features of the end-effectors and condenses them into a set of superpoints.
Subsequently, to capture the global gesture information of each embodiment, 
we apply a geometric transformer \cite{geotransformer} that performs cross-attention over the superpoints.
Directly applying cross-attention over superpoints may lead to 
positional ambiguity. 
To address this issue, we utilize two positional embeddings. The first one is the naive coordinate positional embedding $r^{p}$. Additionally, we design a semantic positional embedding $r^s$,
which provides a unified structural encoding across different embodiments. 
This embedding incorporates spatial semantics to preserve fine-grained 
positional information, thereby enabling more accurate and embodiment-agnostic 
geometry reasoning.

To be specific, we assign each superpoint $p \in \Hat{\mathcal{P}}$ a semantic 2-dimensional index $\pi(p) = (u_p, v_p)$,
where $u_p \in \{0,\ldots,5\}$ denotes the finger-level index (palm: $0$, thumb: $1$, \dots, little finger: $5$) and $v_p \in \mathbb{Z}_{\ge 0}$ denotes the segment-level index along the corresponding finger (each finger's segment levels start from $0$).
We form the 2D index vector $\mathbf{s}_p = [u_p, v_p]^\top \in \mathbb{R}^2$ and map it into the feature space: $r^s = \mathbf{s}_p \mathbf{W}^{S} \in \mathbb{R}^{d_t}$,
where $d_t$ is the superpoint feature dimension, and $\mathbf{W}^{S} \in \mathbb{R}^{2 \times d_t}$ is the projection matrices for semantic embedding.

Then the positional embedding $r^s$ and $r^p$ are added together, and fed into 
the geometric transformer:
\begin{equation}
    \tilde{f}_p = \mathrm{Transformer}\big(f_p, r^p + r^s\big),
    \quad p \in \Hat{\mathcal{P}} .
\end{equation}

Finally, we aggregate the refined superpoint features $\{\tilde{f}_p\}_{p \in \Hat{\mathcal{P}}}$ 
via global average pooling to obtain the GaLR 
$z \in \mathbb{R}^{d_{\mathrm{latent}}}$:
\begin{equation}
    z = \frac{1}{|\Hat{\mathcal{P}}|} 
    \sum_{p \in \Hat{\mathcal{P}}} \tilde{f}_p .
\end{equation}

\noindent
\textbf{Unified decoder with latent retargeting.}
Although GaLR serves as a latent action representation to unify the action dimensions, 
embodiment-specific inference still requires recovering the actual joint angles. 
Prior approaches \cite{latent_action_diffusion} typically train separate action decoders for each embodiment, 
which restricts each decoder to learning only from its own dataset,
preventing effective utilization of shared information, and more importantly, may cause overfitting in few-shot learning scenarios (see \Cref{sec:few-shot learning}).



To address this, we design a unified decoder $g_\psi$ that predicts all joints $\hat{\Theta}$ in a hypothetical universal hand model $\mathcal{H}$, 
which contains every physically meaningful joint across all end-effectors (e.g., thumb yaw, thumb base flexion, index finger yaw, etc.). 
For each specific hand $m$, only the joints that exist on that hand are selected from $\mathcal{H}$, denoted as $\hat{\Theta}^m$.
Finally, we can calculate the RMSE loss of the joint angles $\hat{\Theta}^m$ recovered from GaLR and the ground-truth joint angles $\mathbf{a^m}$, 
allowing the entire GaLR training framework to be optimized in an 
end-to-end manner.

\input{figures/nine_region}

\subsection{Cross-Embodiment Policy Co-Training}

By introducing GaLR as the general latent action representation, we can unify the training and inference process of different end-effectors.
Given the cross-embodiment dataset $\{\mathcal{D}_m\}_{m=1}^M$, 
our objective is to learn a unified visuomotor policy $\pi: \mathcal{O} \to \mathcal{A}$,
that maps visual observations $\mathbf{o} \in \mathcal{O}$ into actions 
$\mathbf{a} \in \mathcal{A}$ across different embodiments. We denote the function that generates GaLR from hand-specific actions (joint angles) as:
\begin{equation}
    \mathcal{G} = f_\theta \circ f_{FK}^m.
\end{equation}
Then, for each embodiment $m \in \mathcal{M}$, the latent policy specializes into 
\begin{equation}
    \pi_m: \mathbf{o}^m_t \mapsto \mathcal{G}( \mathbf{a}^m_t) \in \mathbb{R}^{d_{latent}}.
\end{equation}

In this way, the policy $\pi$ learns embodiment-agnostic visuomotor skills 
from the joint dataset $\{\mathcal{D}_m\}_{m=1}^M$ through GaLR.
OPFA is base-policy-agnostic and can be seamlessly integrated into 
various policy architectures such as ACT \cite{act} or DP3 \cite{dp3}. 
The only modification required is to transform the action prediction 
process into GaLR prediction, while replacing the state terms $\mathbf{s^m_t}$ in the 
observation with the corresponding GaLR representation $\mathcal{G}(\mathbf{s^m_t})$. 
In our implementation, we adopt DP3 as the underlying policy backbone.

During training, the denoising process is applied to GaLR in the latent space. While during inference, we directly use the DP3 model trained on 
$\{\mathcal{D}_m\}^M_{m=1}$ to predict GaLR, and then employ the pretrained 
decoder to recover the embodiment-specific joint angles $\hat{\Theta}^m$.

%% file: figures/nine_region.tex
\begin{figure*}
\centering
\vspace{-6pt}
\includegraphics[width=\textwidth]{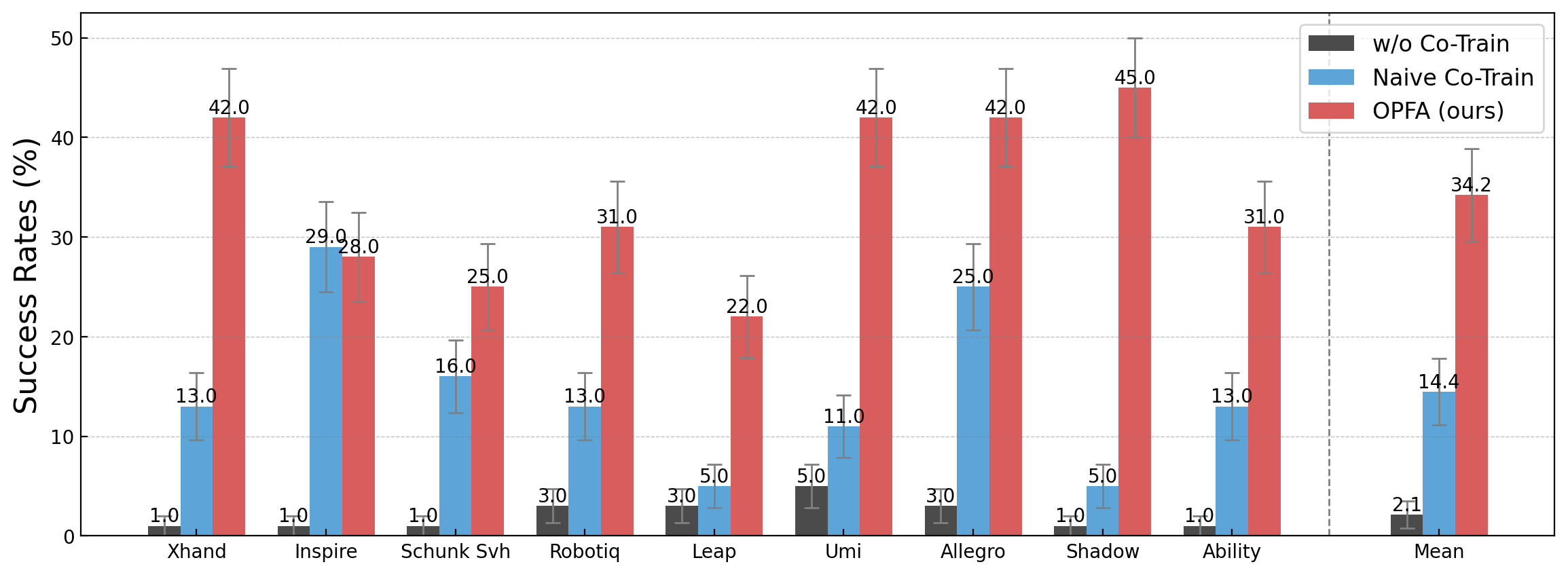}
\vspace{-17pt}
\caption{
\textbf{Spatial generalization evaluation on the spray-picking task.}
Data for each embodiment are collected in distinct regions, and we evaluate the policy for each embodiment to generalize to regions covered by data from the others.
}
\label{fig:nine_region}
\vspace{-12pt}
\end{figure*}

%% file: sec/experiments.tex
\section{EXPERIMENTS}

In studying cross-embodiment manipulation, we typically have two expectations. First, we hope that after co-training across different embodiments,
\textbf{One embodiment can generalize to workspace regions represented in the data of other co-training embodiments.}
Second, when introducing a new embodiment, we aim to \textbf{leverage cross-embodiment data to enable few-shot learning}. To this end, we design comprehensive experiments in this section to validate these two capabilities of OPFA.

\subsection{Implementation Details}

\noindent
\textbf{Embodiments.} Our experiments involve a total of 11 embodiments. In real-world settings, we use XHand \cite{xhand}, Inspire Hand (tactile version) \cite{inspirehand}, Robotiq-2F-85, and Leap Hand \cite{leaphand}, while in simulation, we additionally evaluate on UMI \cite{umi}, Robotiq-3F, Allegro, Shadow Hand, Ability Hand, Schunk SVH Hand, and the non-tactile version of Inspire Hand. All embodiments share the same encoder, decoder, and latent action space.

\noindent
\textbf{Tasks.} In simulation, we evaluate OPFA on seven tasks, including kettle-pulling, button-pressing, bucket-lifting, and pick\&place. In real-world settings, we also conduct experiments on seven tasks.

\noindent
\textbf{Baselines.} To evaluate OPFA’s cross-embodiment capabilities, we compare against two baselines. The first trains only on data from the tested embodiment, denoted as \textit{w/o Co-Train}. The second naively assigns a separate decoder to each embodiment (most current methods use) and performs cross-embodiment co-training, denoted as \textit{Naive Co-Train}.

\input{tables/zero-shot}

\input{figures/few-shot}

\input{figures/nine_few_shot}

\subsection{Generalization across Different Embodiments}

In this section, our goal is to verify that after co-training, each embodiment can generalize within the region covered by data from the other embodiments, or can generalize to similar skills. To this end, we conduct two types of evaluations: spatial generalization and object generalization. In the \textbf{spatial generalization} setup, we collect 72 training trajectories for each embodiment through teleoperation, and the training data of different embodiments are drawn from distinct spatial regions. We then co-train on data from multiple embodiments and evaluate each embodiment within the other embodiments' data distribution regions. \Cref{tab:zero-shot-se} presents the results of Inspire Hand \cite{inspirehand} and XHand \cite{xhand} under these experimental setups.

Since the test regions are entirely unseen for each embodiment, the w/o co-train baseline almost completely fails, exhibiting poor positional generalization across five tasks with both embodiments. In contrast, the naive cross-embodiment co-train baseline partially alleviates this issue, as training on multiple embodiments provides broader wrist-level shared knowledge. However, due to the embodiment gap, transferring the fine-grained end-effector spatial information is extremely challenging, and using separate embodiment-specific decoders exactly discards this information. By comparison, OPFA co-trains multiple embodiments within a shared, geometry-aware latent action space, enabling the joint exploitation of both wrist-level and end-effector-level information. Consequently, OPFA achieves strong generalization even on entirely unseen test data, consistently demonstrating stable cross-embodiment spatial generalization across tasks and attaining success rates above 90\% on tasks such as bucket-lifting and banana-picking.

Moreover, we evaluate OPFA and the baselines on their ability for cross-object skill transfer, which is more challenging. Specifically, we collect data of Inspire Hand grasping a can and XHand grasping a spray, co-train the two datasets, and then test OPFA on cross-object generalization. As shown in \Cref{tab:zero-shot-se}, OPFA significantly outperforms the baselines, achieving a 26\% improvement over naive co-train on Inspire Hand and an 18\% improvement on XHand.

\noindent
\textbf{More Embodiments.} To evaluate the generality of GaLR, we conduct cross-embodiment co-training experiments on nine end-effectors for a spray-picking task. The test region is divided into nine subregions, with each embodiment’s training data drawn from only a single subregion. As shown in Figure 3, under this challenging setup, the w/o co-train baseline completely fails, whereas OPFA significantly improves success rates for all embodiments, achieving an average improvement of 19.8\% over the naive co-train method.

\subsection{Few-Shot Learning Ability}
\label{sec:few-shot learning}

Another key capability of OPFA is few-shot learning—that is, the ability to achieve performance close to single-source, large-data training when a new end-effector is introduced with only a small amount of data. To evaluate this, we conduct experiments on the banana-picking task, collecting 72 trajectories each for the Inspire Hand and XHand. Subsets of these trajectories are then sampled for few-shot learning tests. For the Inspire Hand, subsets of 1, 2, 4, 8, 12, 18, and 36 trajectories are co-trained with all 72 trajectories of XHand, and vice versa for XHand. The resulting few-shot learning curves are shown in Figure 4. OPFA’s performance improves rapidly with the number of sampled trajectories: with only 8 trajectories, the Inspire Hand already achieves a high success rate, while XHand surpasses 80\%, approaching the performance of fully trained models and far exceeding baseline methods. In contrast, naive co-training suffers from training inhibition as the sample size increases, sometimes performing worse than the w/o co-train baseline. These results demonstrate OPFA’s strong few-shot learning capability.

\noindent
\textbf{More Embodiments.} As shown in \Cref{fig:few-shot}, OPFA substantially enhances cross-embodiment few-shot learning, achieving competitive success rates with as few as eight newly collected trajectories. To further validate OPFA’s few-shot capability across a wide range of end-effectors, we expand the number of embodiments to nine. In the spray-picking task, only eight trajectories are collected for each embodiment, and the combined set of 72 trajectories is used for co-training. The results, presented in \Cref{fig:nine_few_shot}, reveal that when the number of embodiments is large, the naive co-train method may suffer from conflicts due to the significant geometric gaps between different grippers and dexterous hands, in some cases even underperforming the w/o co-train baseline.
In contrast, OPFA leverages geometry-aware co-training to effectively integrate information across embodiments, leading to substantial accuracy improvements—for example, achieving a 93\% success rate on Leap Hand and an average success rate of 62.1\%, which represents a gain of over 20\% compared to the naive co-train method.

\input{figures/real_expe_tmp}

\input{tables/realworld}

\subsection{Real World Experiments}

\noindent
\textbf{Experimental Setup.}
Our real-world setup consists of a UR5e arm and a Microsoft Azure Kinect depth camera (\Cref{fig:asset}), equipped with four different end-effectors: XHand, Inspire Hand, Leap Hand, and the Robotiq-2F gripper.
We design seven manipulation tasks involving diverse objects and objectives: \textbf{Basket}, pick and place a basket; \textbf{Tissue}, pull a tissue from a dispenser; \textbf{Broom}, sweep an object into a dustpan with a broom; \textbf{Pot}, grasp a pot and pour into a cup; \textbf{Mango}, pick and place a mango onto a plate; \textbf{Drawer}, place a block into a drawer and close it; \textbf{Syringe}, position a syringe over a cup and press the plunger.

For each embodiment and task, 24 demonstrations were collected via teleoperation using a Spacemouse \cite{spacemouse} and an exoskeleton device \cite{homie}. During evaluation, \textbf{one unified policy checkpoint} is tested across all embodiments, with 10 trials per task. Due to morphological limitations, some embodiments were excluded from certain tasks (e.g., Robotiq-2F from \textit{Syringe}/\textit{Pot}, Inspire Hand from \textit{Syringe}).

\input{figures/asset}

\noindent
\textbf{Results.} The quantitative results of our real-world experiments are summarized in \Cref{tab:realworld}. Our method demonstrates consistently high success rates across most of the evaluated tasks and embodiments, validating its capability to learn a versatile and generalizable manipulation policy.
Notably, on challenging long-horizon tasks such as \textit{Drawer} and tasks requiring fine-grained control such as \textit{Pot}, methods trained without co-training can only learn conceptual knowledge from single-embodiment data, resulting in large operational errors when object positions vary widely. Naive co-training improves performance on long-horizon tasks to some extent, but on the \textit{Pot} task, substantial structural differences across end-effectors induce action conflicts, leading to worse performance than w/o co-train. In contrast, OPFA enables shared learning of both wrist-level trajectory concepts and fine-grained actions across embodiments, consistently outperforming baseline methods.
By sharing geometric knowledge across different end-effectors, \textbf{OPFA} achieves a level of performance that embodiment-specific heads cannot, especially in data-efficient settings.
Moreover, OPFA's skill transfer capability is task-agnostic: it can generalize across both deformable (Tissue) and rigid (Broom) objects, and across both pick\&place (Mango) and dexterous (Syringe) tasks.

%% file: tables/zero-shot.tex
\begin{table}
  \centering
  \caption{
  \textbf{Zero-shot skill transfer experiments.} In the table, \textit{Position Generalization} refers to different data collection regions for different embodiments, while \textit{Object Generalization} refers to different manipulation objects.
  }
  \resizebox{\columnwidth}{!}{\begin{tabular}{l|ccccc|c}
    \toprule
    Task & Kettle & Button & Sanitizer & Bucket & Banana & \textbf{Pick Spray\&Can} \\
    \midrule
    & \multicolumn{5}{c|}{Spatial Generalization} & Object Generalization \\
    \midrule
    \multicolumn{7}{c}{Inspire Hand Success Rate (\%)} \\
    \midrule
    w/o Co-training & 30.0 & 26.0 & 5.0 & 3.0 & 10.0 & 1.0 \\
    Naive Co-training & 57.0 & 39.0 & 71.0 & 50.0 & 83.0 & 57.0 \\
    OPFA (ours) & \textbf{83.0} & \textbf{60.0} & \textbf{82.0} & \textbf{75.0} & \textbf{98.0} & \textbf{83.0} \\
    \midrule
    \multicolumn{7}{c}{Xhand Success Rate (\%)} \\
    \midrule
    w/o Co-training & 3.0 & 11.0 & 4.0 & 1.0 & 5.0 & 41.0 \\
    Naive Co-training & 5.0 & 38.0 & \textbf{61.0} & 33.0 & 30.0 & 53.0 \\
    OPFA (ours) & \textbf{7.0} & \textbf{75.0} & 51.0 & \textbf{94.0} & \textbf{67.0} & \textbf{71.0} \\
    \bottomrule
  \end{tabular}}
  \label{tab:zero-shot-se}
  \vspace{-8pt}
\end{table}


%% file: figures/few-shot.tex
\begin{figure*}
\centering
\vspace{-1mm}
\begin{subfigure}[b]{0.99\columnwidth}
    \centering
    \includegraphics[width=\columnwidth]{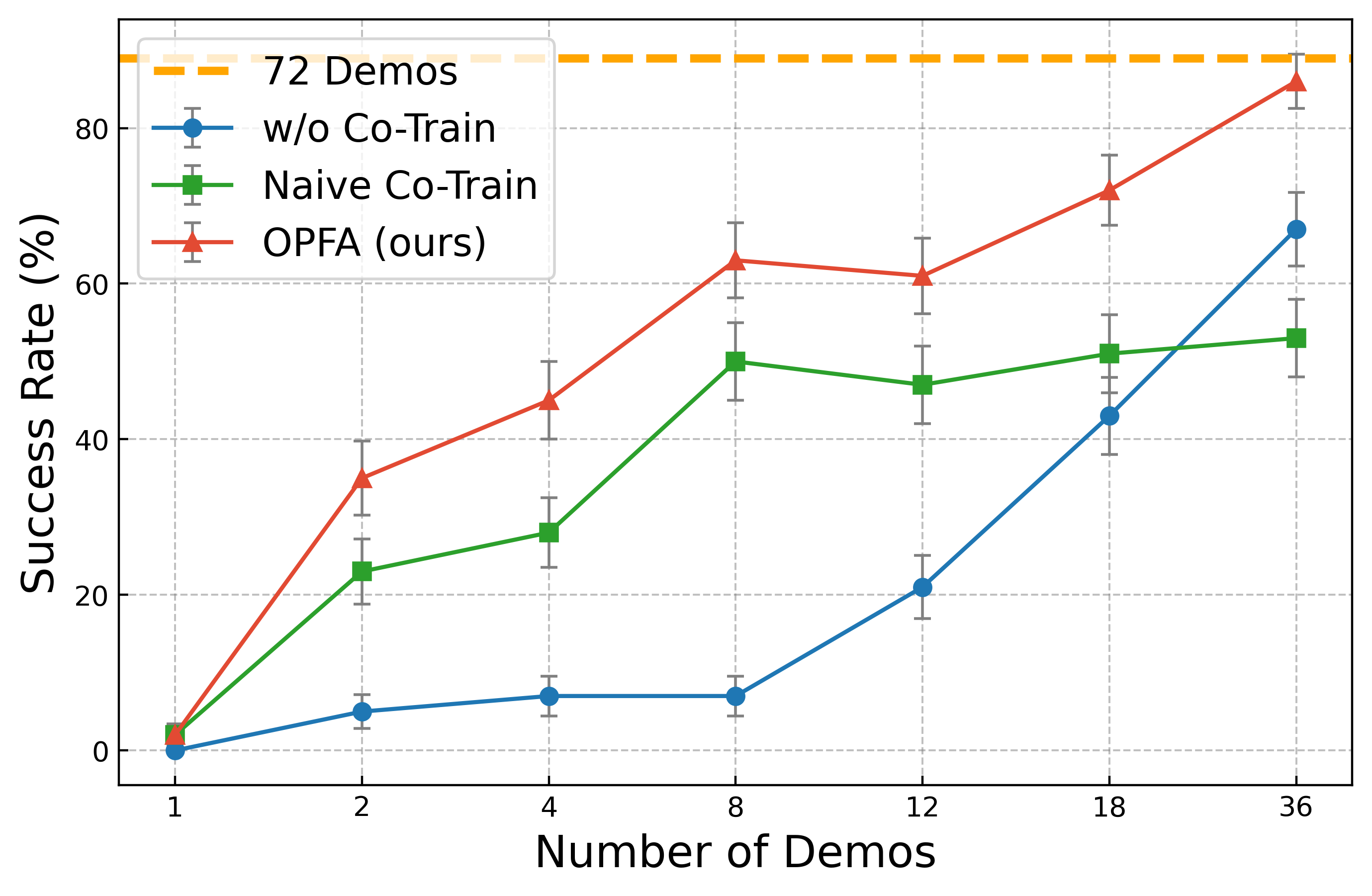}
\end{subfigure}
\hfill
\begin{subfigure}[b]{0.99\columnwidth}
    \centering
    \includegraphics[width=\columnwidth]{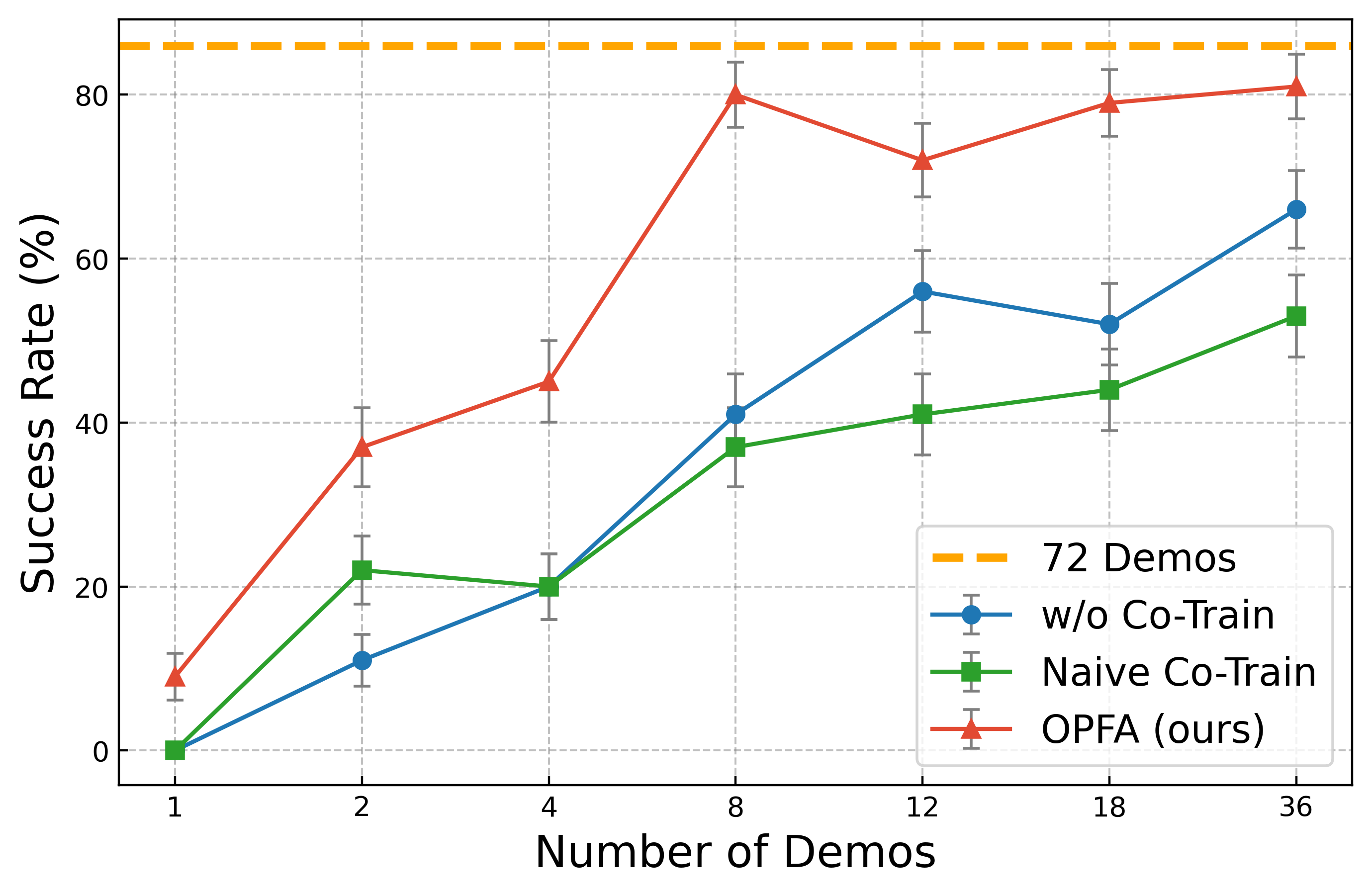}
\end{subfigure}
\vspace{-6pt}
\caption{\textbf{Few-shot learning curves with different demo numbers on the banana-picking task of \textbf{(Left)} Inspire Hand and \textbf{(Right)} XHand}.
Each embodiment is co-trained with 72 demonstrations from the other.
}
\label{fig:few-shot}
\vspace{-8pt}
\end{figure*}

%% file: figures/nine_few_shot.tex
\begin{figure*}
\centering
\includegraphics[width=\textwidth]{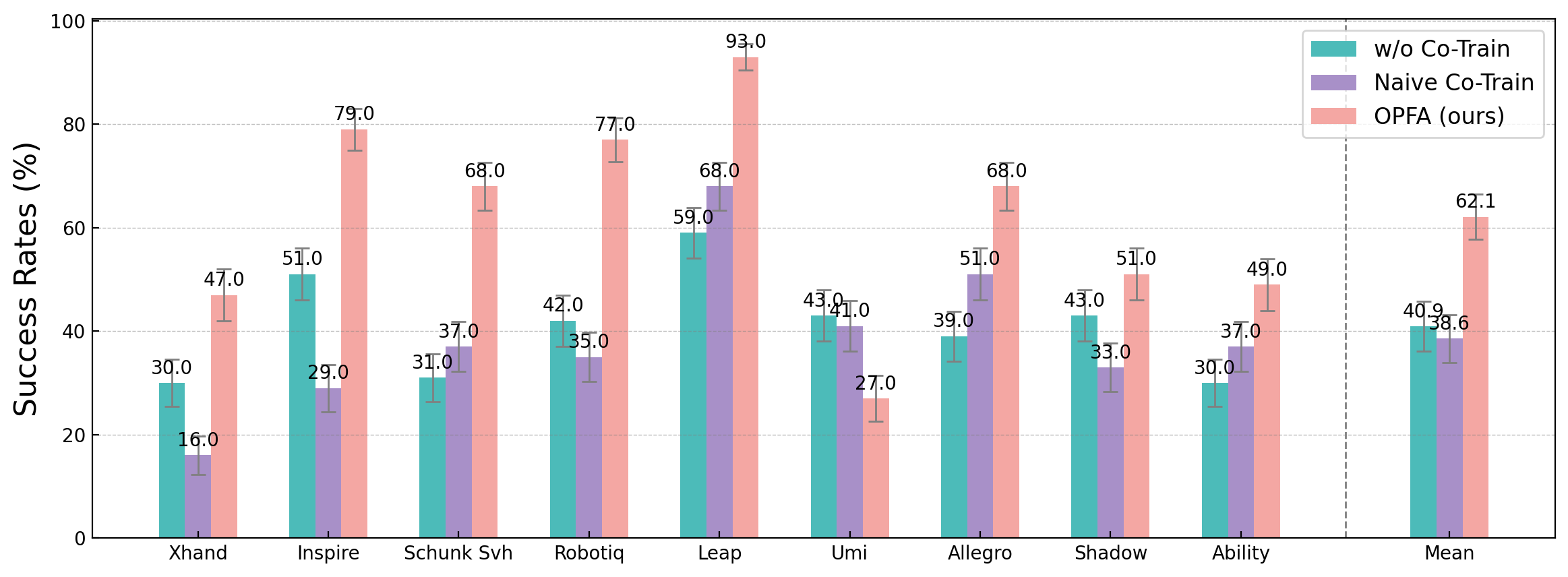}
\vspace{-15pt}
\caption{\textbf{Few-shot learning performance across nine different embodiments on the spray-picking task.} 
We collect eight trajectories for each of the nine end-effectors and co-train on the full dataset to evaluate the few-shot learning capability for each embodiment. OPFA yields a 20\%+ average performance gain over the baselines.
}
\label{fig:nine_few_shot}
\vspace{-10pt}
\end{figure*}

%% file: figures/real_expe_tmp.tex
\begin{figure*}
\centering
\vspace{-1mm}
\includegraphics[width=\textwidth]{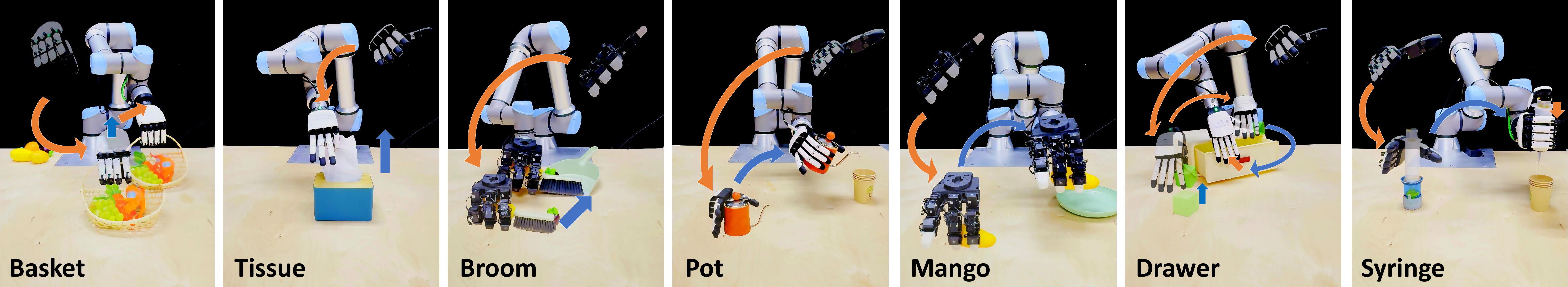}
\vspace{-6mm}
\caption{Tasks for real world evaluation. \textit{Basket} and \textit{Mango} are one-stage pick-and-place tasks.
\textit{Tissue} and \textit{Pot} are contact-intensive tasks, involving deformable object handling or precise manipulation.
\textit{Broom} and \textit{Drawer} are long-horizon sequential tasks involving tool-use and multi-step coordination.
\textit{Syringe} is a fine-motor control task demanding precise force application.}
\label{fig:realEx}
\vspace{-6pt}
\end{figure*}

%% file: tables/realworld.tex
\begin{table}
\vspace{-2mm}
  \centering
  \caption{Real-world experiments success rates. Best and second-best methods are highlighted with different colors.}
  \resizebox{\columnwidth}{!}{\begin{tabular}{l|ccccccc}
    \toprule
    Task & Basket & Tissue & Broom & Pot & Mango & Drawer & Syringe \\
    \midrule
    \multicolumn{8}{c}{Inspire Hand Success Rate (\%)}\\
    \midrule
    w/o Co-training & \second{90} & \best{\textbf{70}} & 70 & \second{80} & 60 & 60 & - \\
    Naive Co-training & \second{90} & \best{\textbf{70}} & \second{80} & 70 & \second{80} & \second{70} & - \\
    OPFA (ours) & \best{\textbf{100}} & 60 & \best{\textbf{100}} & \best{\textbf{100}} & \best{\textbf{100}} & \best{\textbf{80}} & - \\
    \midrule

    \multicolumn{8}{c}{Leap Hand Success Rate (\%)}\\
    \midrule
    w/o Co-training & \second{90} & \second{70} & \second{80} & \second{40} & \second{60} & \second{70} & 60 \\
    Naive Co-training & 80 & 50 & \best{\textbf{90}} & 30 & 10 & 60 & \second{70} \\
    OPFA (ours) & \best{\textbf{100}} & \best{\textbf{90}} & \best{\textbf{90}} & \best{\textbf{80}} & \best{\textbf{90}} & \best{\textbf{90}} & \best{\textbf{100}} \\
    \midrule

    \multicolumn{8}{c}{Robotiq-2F Success Rate (\%)}\\
    \midrule
    w/o Co-training & 30 & \second{90} & - & \second{50} & \second{70} & 60 & - \\
    Naive Co-training & \best{\textbf{90}} & 70 & - & 30 & \second{70} & \second{80} & - \\
    OPFA (ours) & \best{\textbf{90}} & \best{\textbf{100}} & - & \best{\textbf{60}} & \best{\textbf{80}} & \best{\textbf{100}} & - \\
    \midrule
    
    \multicolumn{8}{c}{Xhand Success Rate (\%)}\\
    \midrule
    w/o Co-training & \second{80} & \second{80} & 70 & \second{70} & \second{60} & 50 & 60 \\
    Naive Co-training & \second{80} & 70 & \second{80} & 60 & \second{60} & \second{60} & \second{70} \\
    OPFA (ours) & \best{\textbf{90}} & \best{\textbf{100}} & \best{\textbf{100}} & \best{\textbf{90}} & \best{\textbf{100}} & \best{\textbf{90}} & \best{\textbf{100}} \\
    \bottomrule
  \end{tabular}}
  \label{tab:realworld}
  \vspace{-4mm}
\end{table}

%% file: figures/asset.tex
\begin{figure}
\vspace{-2mm}
\centering
\includegraphics[width=\columnwidth]{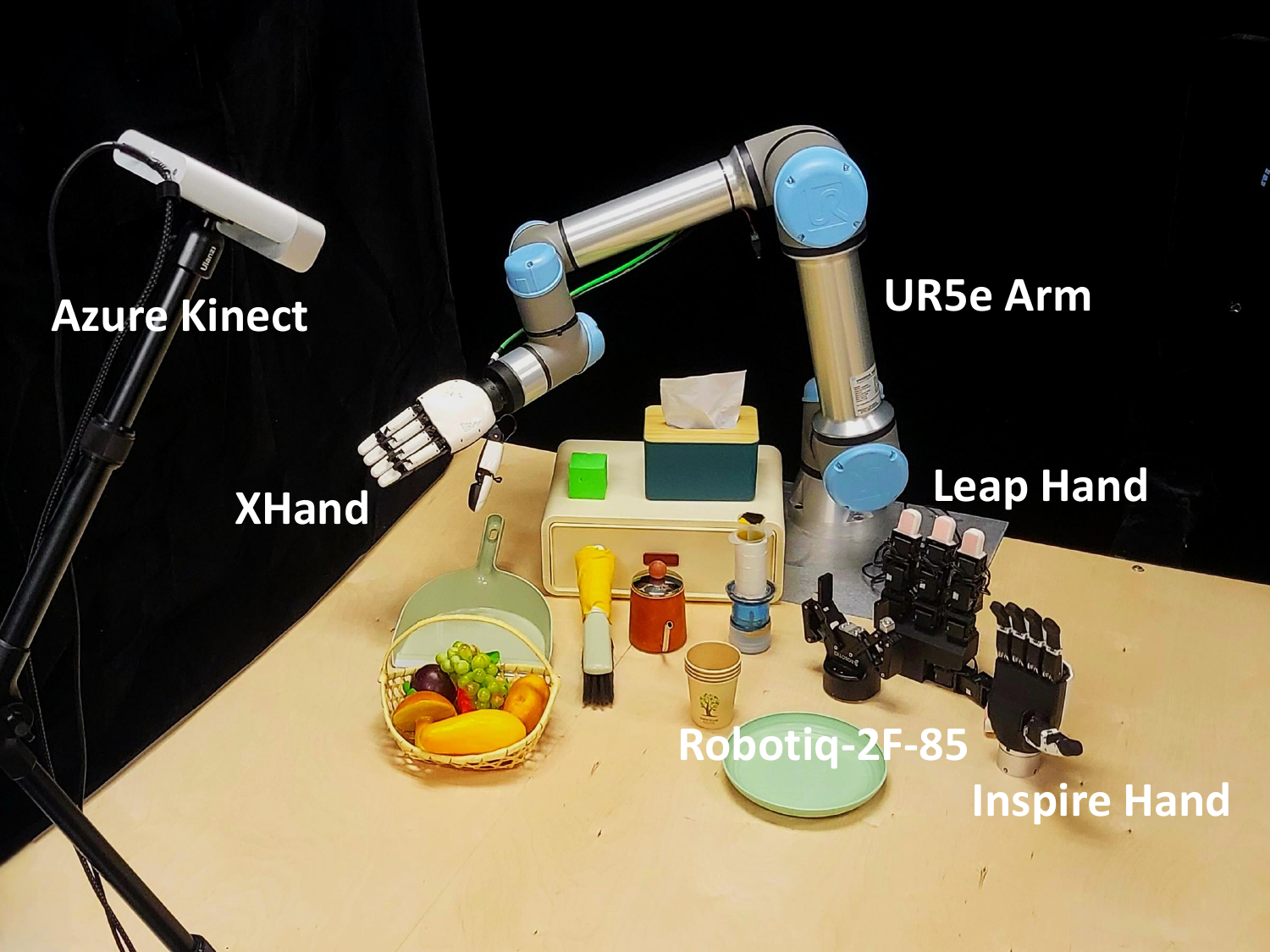}
\caption{Real world experiment setup.}
\label{fig:asset}
\vspace{-4mm}
\end{figure}

%% file: sec/conclusion.tex
\section{CONCLUSIONS}

We introduce One-Policy-Fits-All (OPFA), a unified framework for cross-embodiment manipulation that enables end-to-end co-training across data from diverse robotic hands and grippers. OPFA first learns a Geometry-Aware Latent Representation (GaLR), which constructs a shared latent action space to capture commonalities across different embodiments. It then employs a unified latent retargeting decoder to map this latent representation into embodiment-specific actions without any per-embodiment tuning. Extensive experiments on 11 end-effectors show that OPFA substantially enhances skill transfer and data efficiency, significantly improving the performance of each embodiment while exhibiting strong few-shot learning capabilities.

\noindent
\textbf{Acknowledgements}. This work is supported by Shanghai Artificial Intelligence Laboratory.